\documentclass[sigconf]{acmart}

\AtBeginDocument{%
  }

\setcopyright{acmlicensed}
\copyrightyear{2026}
\acmYear{2026}
\acmDOI{XXXXXXX.XXXXXXX}
\acmConference[DAC '26]{Make sure to enter the correct
  conference title from your rights confirmation email}{June 03--05,
  2018}{Woodstock, NY}
\acmISBN{978-1-4503-XXXX-X/2018/06}
\usepackage{tcolorbox}
\usepackage{xcolor} % Useful for defining custom colors
\usepackage{booktabs}
\usepackage{amsmath} % For using text within math mode if needed
\usepackage[utf8]{inputenc}
\usepackage{tcolorbox}
\tcbuselibrary{skins}

\usepackage{tikz} % Ensure this is loaded

\newcommand{\takeawaybox}[2]{%
    \begin{tcolorbox}[
        boxrule=0pt,        
        leftrule=2pt,       % This creates the left border
        colframe=black,     % Color of the left border
        arc=0pt,            % Sharp corners so the border looks like a straight line
        boxsep=2pt,         
        colback=gray!5!white, 
        enlarge left by=2pt, 
        width=\linewidth,
        top=0pt, 
        bottom=0pt,
    ]\parbox{\linewidth}{
        \setlength{\parskip}{0pt}% 
        \setlength{\parindent}{0pt}% 
        \textbf{Takeaway #1.} \textit{#2}%
    }
    \end{tcolorbox}%
}
\usepackage{multirow}
\usepackage{makecell}
\usepackage{array}
\usepackage[table]{xcolor} % Required for \cellcolor

\begin{document}

% --- Custom Commands (Cleaned up duplicates) ---
\newcommand{\mypar}[1]{
\par\addvspace{0\baselineskip}%
\noindent{\textbf{#1.}}}
\newcommand{\takeaway}[2]{~\\\noindent{$\triangleright$~\textit{Takeaway \##1: #2.}}}
\newcounter{obs}

\newcommand{\observation}[1]{%
    \stepcounter{obs}%
    \par\addvspace{0.2\baselineskip}% 
    \noindent 
    {$\triangleright$~\textit{\bf \underline{\em Observation \#\arabic{obs}.}} {\em #1: }}
}

%%
%% The "title" command
\title{An experimental study of KV cache reuse strategies in chunk-level caching systems}

% --- Authors (Horizontal Layout with symbols and line breaks) ---
% The \\ command forces a new line within the author list.
% --- Authors (Horizontal Layout with symbols and line breaks) ---
\author{%
  Samuel Cestola$^{\ast, \dag}$, Tianxiang Xia$^{\ast, \dag}$, Zheng Weiyan$^{\dag}$, \\
  Zheng Pengfei$^{\ddag}$, Diego Didona$^{\dag}$%
}

% Empty affiliation block so ACM doesn't throw a compilation error,
% keeping the space under the authors completely clean.
\affiliation{%
  \institution{}
  \country{}
}

% All legends moved to the bottom as clean footnotes
\thanks{$^{\ast}$Equal contribution.}
\thanks{$^{\dag}$Huawei, Switzerland.}
\thanks{$^{\ddag}$Huawei, China.}

\renewcommand{\shortauthors}{Cestola and Xia, et al.}

% --- Abstract (Removed the duplicate one) ---
\begin{abstract}
Retrieval-augmented generation improves large language models' accuracy by adding relevant retrieved text to the prompt. Chunk-level caching (CLC) accelerates inference by precomputing KV caches for these retrieved chunks and reusing them. However, these caches miss cross-attention dependencies between chunks, which can reduce output quality. Several methods try to improve CLC accuracy using different techniques. We make two main contributions.
First, we show that existing CLC approaches have {\em fundamental} limitations that limit their accuracy or their applicability. We back this conclusion with an extensive CLC system experimental evaluation. Second, we observe that existing CLC techniques are complementary. We leverage this insight to propose a new CLC design that carefully combines them and achieves better accuracy.
\end{abstract}

\settopmatter {printacmref=false}
\renewcommand\footnotetextcopyrightpermission[1]{}

% --- CCS & Keywords (Removed the duplicates) ---
\begin{CCSXML}
<ccs2012>
  <concept>
    <concept_id>10010147.10010178.10010179</concept_id>
    <concept_desc>Computing methodologies~Language models</concept_desc>
    <concept_significance>500</concept_significance>
  </concept>
</ccs2012>
\end{CCSXML}

\ccsdesc[500]{Computing methodologies~Language models}

\keywords{RAG, LLM, chunk-level caching}

% --- Maketitle (Must be here, only once) ---
\maketitle

\section{Introduction}
Retrieval-augmented generation (RAG) is a widely used strategy to improve the comprehensiveness and factual reliability, hereafter, {\em accuracy}, of large language models (LLM) ~\cite{lewis2021retrievalaugmentedgenerationknowledgeintensivenlp,metis,hydrarag}. RAG enriches user queries with online-retrieved text chunks, supplying up-to-date and domain-specific information that reduces hallucinations. While effective, RAG substantially increases inference latency because it inflates the prompt length. This problem is amplified by the fact that multi-head attention, the mechanism at the core of the Transformer architecture, incurs a quadratic computational overhead with respect to the prompt length to compute the corresponding attention states, the so called Key-Value (KV) caches~\cite{vaswani2023attentionneed}.

%not sure about the exact papers in this citation here
Prefix caching is a common method to mitigate the high latency incurred with long prompts~\cite{zheng2024sglangefficientexecutionstructured, qin2025mooncakekvcachecentricdisaggregatedarchitecture,ragboost}. Prefix caching stores the KV caches obtained for user queries and reuses them when a new request begins with the same initial token sequence. The attention states of each token, however, depend on all preceding tokens. % the KV cache  of a prompt P is reusable for a prompt P'  only if  P  is an exact prefix of P'. 
Prefix caching is therefore effective only when prompts share identical preceding chunks,  but it  can suffer from limited reuse as the same RAG chunks often appear at different positions~\cite{agarwal2025cachecraftmanagingchunkcachesefficient,jiang2025ragboostefficientretrievalaugmentedgeneration}.

Chunk-level caching (CLC) addresses this limitation by computing the KV caches of individual chunks independently and pursues position-independent reuse to avoid redundant computation. Direct reuse at arbitrary positions, however, causes severe accuracy degradation for two main reasons.\footnote{\scriptsize A third issue arises from positional encoding, e.g., RoPE ~\cite{su2023roformerenhancedtransformerrotary}, because precomputed caches all start at position zero. This is handled similarly across systems by rotating the cached KV tensors to the target positions, and hence we do not discuss this aspect.}
First, in standard LLM inference, each KV entry reflects {\em cross-chunk} attention from all preceding tokens. Independently computed chunk caches include only {\em intra-chunk} attention states and miss semantic dependencies between tokens across chunks.
Second, attention patterns, such as strong focus on early tokens and locality biases~\cite{hu2025epicefficientpositionindependentcaching,yang2025apefasterlongercontextaugmented}, behave differently when processing a single contiguous prompt versus isolated chunks.

\begin{figure}[t!]
    \centering
    \includegraphics[scale=.3]{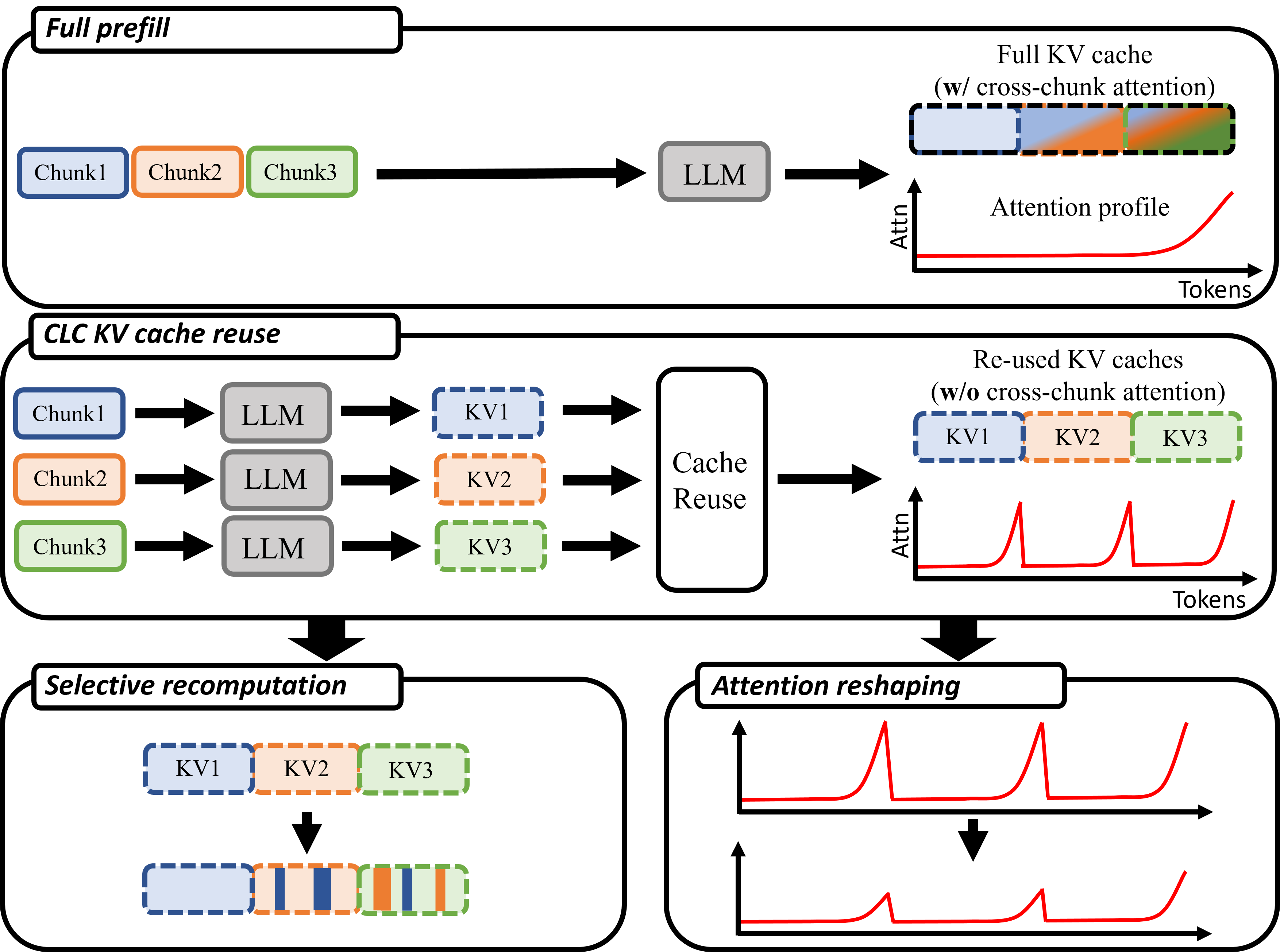}\
  \caption{Top: KV cache and attention profile with full-prefill. Middle: CLC with naive KV cache reuse lacks cross-chunk attention, and obtains different attention profiles. Bottom: selective recomputation  aims to recover key cross-chunk attention information; attention reshaping  aims to obtain an attention profile similar to full prefill's.}\label{fig:intro}
\end{figure}

Existing CLC solutions to address these challenges fall into two categories: recomputation-based methods, which selectively recompute attention for a subset of tokens to restore missing cross-chunk information; and attention-reshaping methods, which modify attention to approximate the behavior of non-cached inference.
\mypar{Contributions} We perform an extensive design analysis and experimental evaluation of existing CLC designs. Our study allows us to make two contributions in the field of CLC  design. First, we show that all existing approaches exhibit {\em fundamental} limitations that affect their accuracy and applicability (Section~\ref{sec:accuracy}). Second, we observe that different solutions targeting different aspects of CLC can be leveraged synergistically. We materialize this observation into a new CLC design that combines existing approaches and achieves up to 5\% higher accuracy than the single approaches taken singularly (Section~\ref{sec:golem}). We also discuss extensions to our work, and future research avenues (Section~\ref{sec:conclusion}). %achieves up to 6\% higher accuracy than state-of-the-art ones (Section~\ref{sec:golem}). We also discuss limitations of our work, and future research avenues (Section~\ref{sec:conclusion}).

\begin{table}[t!]
\scriptsize

\begin{tabular}{|>{\centering\arraybackslash}m{1.5cm}|>{\centering\arraybackslash}m{1.6cm}|>{\centering\arraybackslash}m{4.5cm}|}
\hline
\textbf{Taxonomy} & \textbf{System} & \textbf{Technique} \\
\hline
\begin{tabular}{@{}c@{}} Selective \\ Recomputation \end{tabular} 
& Cacheblend ~\cite{yao2025cacheblendfastlargelanguage} & Recompute KV cache entries with highest deviation from full-prefill \\
\cline{2-3}
& Epic ~\cite{hu2025epicefficientpositionindependentcaching} & Recompute beginning-of-chunk KV cache entries \\
\cline{2-3}
& Cacheclip ~\cite{yang2025cacheclipacceleratingrageffective} & Choose recomputation tokens at last layer of auxiliary model \\
\cline{2-3}
& Droidspeak ~\cite{liu2025droidspeakkvcachesharing} & Re-evaluate chosen tokens at mid layer \\
\hline
\begin{tabular}{@{}c@{}} Reshape attention: \\ modify attn.\ scores \end{tabular}
& Link0 ~\cite{hu2025epicefficientpositionindependentcaching} & Absorbs abnormally high attention to beginning-of-chunk-tokens \\
\cline{2-3}
& APE ~\cite{yang2025apefasterlongercontextaugmented} & Rescales attention to resemble full-prefill one \\
\cline{2-3}
& SEL ~\cite{zhang2025attentionentropykeyfactor} & Reduce entropy: query attends subset of chunks. \\
\hline
\begin{tabular}{@{}c@{}} Reshape attention: \\ fine-tuning \end{tabular}
& TurboRAG ~\cite{lu2024turboragacceleratingretrievalaugmentedgeneration} & Fine-tune attention w/o cross-chunk attention \\
\cline{2-3}
& BlockAttention ~\cite{ma2025blockattentionefficientprefilling} & Fine-tune attention w/o cross-chunk attention \\
\cline{2-3}
& KVLink ~\cite{yang2025kvlinkacceleratinglargelanguage} & As above, plus adding extra tokens with full attention \\
\hline
\end{tabular}
\centering
\caption{State-of-the-art CLC approaches.}\label{tab:systems}
\end{table}

\section{Background}
%assume that by attention is attention is all you need
Most LLMs are built upon the Transformer architecture, which is realized through a stack of identical computational layers~\cite{vaswani2023attentionneed}. Within each layer, the input tensor (activations) is first linearly projected into three distinct matrices: Query ($\mathbf{Q}$), Key ($\mathbf{K}$), and Value ($\mathbf{V}$). The core operation is the self-attention mechanism, which captures the contextual dependency of a token (via $\mathbf{Q}$) on all preceding tokens (via $\mathbf{K}$ and $\mathbf{V}$) in the sequence. The scaled dot-product attention output ($\mathbf{A}$) is computed as $\mathbf{A}(\mathbf{Q}, \mathbf{K}, \mathbf{V}) = \text{Softmax}\left(\mathbf{Q}\mathbf{K}^{\text{T}}/\sqrt{d_k}\right)\mathbf{V}$, where $d_k$ is the dimension of the model. This attention block is replicated, within each layer, across $H$ independent heads. %, each tasked with learning a specialized set of relational weights. 
The resultant outputs are concatenated and channeled through a neural network, which produces the activations for the subsequent layer. 

LLMs operate in two phases: prefill, which processes the initial prompt, and decode, which produces the answer one token at a time. The  prefill phase processes the entire input prompt  at once, resulting in the attention calculation having quadratic computational complexity with respect to the input sequence length. To mitigate this cost, the computed $\mathbf{K}$ and $\mathbf{V}$ matrices are stored in the KV cache during prefill. This is critical for efficiency, as it allows the subsequent  decoding phase to reuse the K and V matrices.

RAG systems augment the prompt with extra chunks of text that are relevant to the target  query to improve the accuracy of the output. To avoid recomputing the KV matrices for the same text chunks across different queries, CLC systems pre-compute and store the KV caches of the text chunks. This is executed by performing a prefill operation on the text chunks to generate chunk-specific $\mathbf{K}$ and $\mathbf{V}$ matrices, which  are then stored for later retrieval.  Figure~\ref{fig:intro} compares an LLM query without KV cache reuse, hereafter referred to as {\em full prefill}, with a query with KV cache reuses in a CLC system.
CLC enables higher efficiency, but it is fundamentally limited: the local prefill operation constrains the attention span of each chunk to its internal tokens, leading to an absence of cross-chunk attention in the resulting $\mathbf{KV}$ caches, which constitutes a significant challenge to maintaining high  accuracy.
\section{Methodology}\label{sec:methodology}
In this section we present the models, datasets, and metrics we use.
\mypar{Models} We use Llama3.1-8B-Instruct ~\cite{grattafiori2024llama3herdmodels}, Qwen3-8B ~\cite{yang2025qwen3technicalreport}, and Mistral-7B-Instruct-v0.3 ~\cite{jiang2023mistral7b}, which are widely used to evaluate CLC systems. We focus on 7B/8B models for two reasons: $i)$ among the top 10 most downloaded models from the three model families,  more than $85\%$ of the downloads are for 7B/8B variants~\footnote{Data retrieved on Oct 27th 2025 on huggingface.co}; $ii)$ small models tend to be more parameter-efficient, i.e., they have less redundancy than large models, making it harder to improve performance with approximation techniques without compromising accuracy ~\cite{smallvslarge}.
\mypar{Datasets} We focus on multi-hop reasoning Q\&A datasets for two reasons: $i)$ answering correctly requires the LLM to reason about multiple chunks in the query, which is crucial to analyze cross-chunk attention dynamics; and $ii)$ evaluating the accuracy of the answer is easier than in other tasks, such as summarization.  We use the popular 2WikiMQA ~\cite{ho2020constructingmultihopqadataset} and  Musique ~\cite{trivedi2022musiquemultihopquestionssinglehop} datasets, used in  the original evaluations of 9 out of the 11 systems we consider, and the RULER ~\cite{hsieh2024rulerwhatsrealcontext} dataset, which is used by the remaining two~\cite{yang2025cacheclipacceleratingrageffective,zhang2025attentionentropykeyfactor}.

\mypar{Evaluation metric} We focus on evaluating the accuracy of CLCs. We use the $\mathbf{F1}$ score to measure the similarity between model output and ground-truth answers~\cite{f1}, a standard practice in CLC literature. 
We note that prior evaluations aggregate the F1 score over all queries, even when the baseline model with full prefill fails. In our experiments, the baseline models achieve F1=0 in 42–58\% of queries. This masks the real impact of KV cache reuse: both the baseline and approximated models score F1=0 on these hard queries, producing no measured difference and overstating the technique’s effectiveness. We therefore use an adjusted F1 that includes only queries where the baseline achieves a non-zero score, better reflecting how well a method preserves the model’s inherent accuracy.
\mypar{Systems} We integrate the systems we evaluate experimentally in the same framework based on HuggingFace Transformers library. For systems whose source code is available~\cite{yao2025cacheblendfastlargelanguage,yang2025apefasterlongercontextaugmented}, we re-use as much code as possible; we implement from scratch the ones that are closed-source~\cite{liu2025droidspeakkvcachesharing,hu2025epicefficientpositionindependentcaching,yang2025cacheclipacceleratingrageffective}, or not easily integrated with our framework~\cite{ma2025blockattentionefficientprefilling}, doing our best to implement their salient features.

\section{Experimental study}\label{sec:accuracy}
In this section we present, discuss, and evaluate different approaches implemented by state-of-the-art systems to improve the accuracy of KV cache reuse in CLC. Table~\ref{tab:systems} describes the systems we discuss. 
\subsection{Recomputation-based approaches}
In this section we discuss the CLC designs  that recompute the cross-chunk attention for a subset of tokens.
\mypar{Cacheblend~\cite{yao2025cacheblendfastlargelanguage}} The main insight in Cacheblend  is that recomputing the KV caches with cross-chunk attention  for a subset of the tokens in the re-used KV caches can approximate the accuracy of a full prefill. To identify the critical tokens for the recomputation, Cacheblend runs a normal prefill at layer one of the LLM. The K matrices obtained starting from the output of layer one are  compared  with the cached ones to compute a per-token difference metric, hereby referred to as $\Delta K$. For a token $t$ and for $H$ heads, $\Delta K[t] = \sum_{i=1}^H ||K^i_p[t] - K^i_c[t]||_2^2$, where $c$ denote the cached matrices, and $p$ the ones resulting from the layer-1 prefill.  The $r\%$ tokens that exhibit the highest $\Delta K$  are selected for recomputation, meaning that only the KV cache entries of such tokens are  recomputed in the prefill of the remaining layers.

\begin{figure}[tbp]
    \centering
    \includegraphics[width=\linewidth]{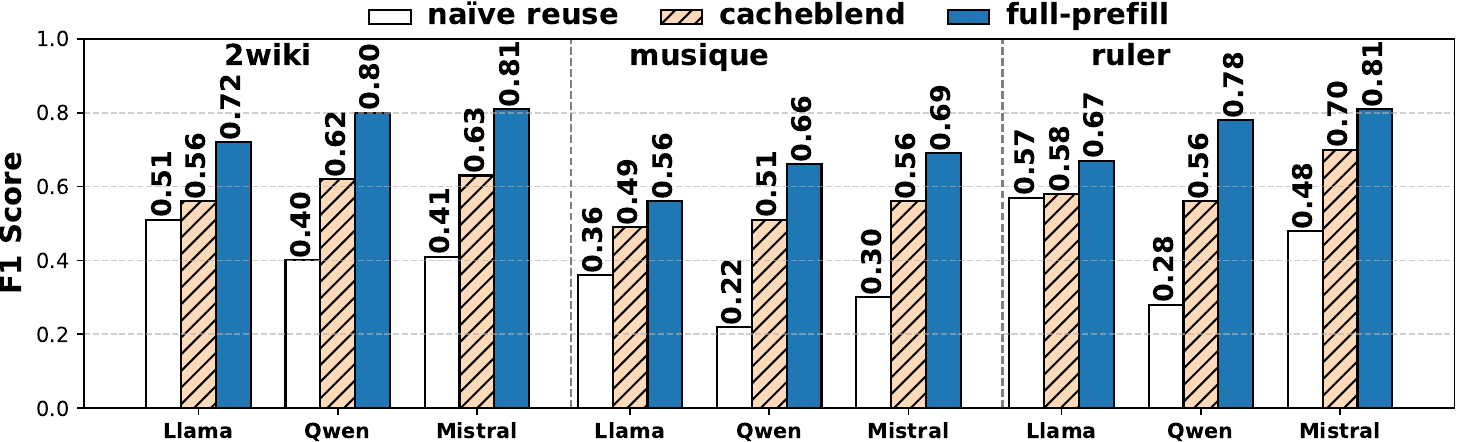}
  \caption{Cacheblend accuracy.}\label{fig:cacheblend}
\end{figure}

Figure~\ref{fig:cacheblend} shows the accuracy achieved by Cacheblend using $r=15\%$ (a value used in many CLC systems as default) and compares it with the accuracy achieved by full prefill and by naive reuse, i.e., $r=0$, which is the approach implemented by PromptCache, to the best of our knowledge the earliest KV cache reuse system~\cite{gim2024promptcachemodularattention}. Cacheblend outperforms naive reuse, but it is also substantially less accurate than full prefill, with differences ranging from 7\% to 18\%. 

\begin{figure}[b]
  \centering
  \begin{minipage}[b]{0.48\linewidth}
    \centering
    \includegraphics[width=\linewidth]{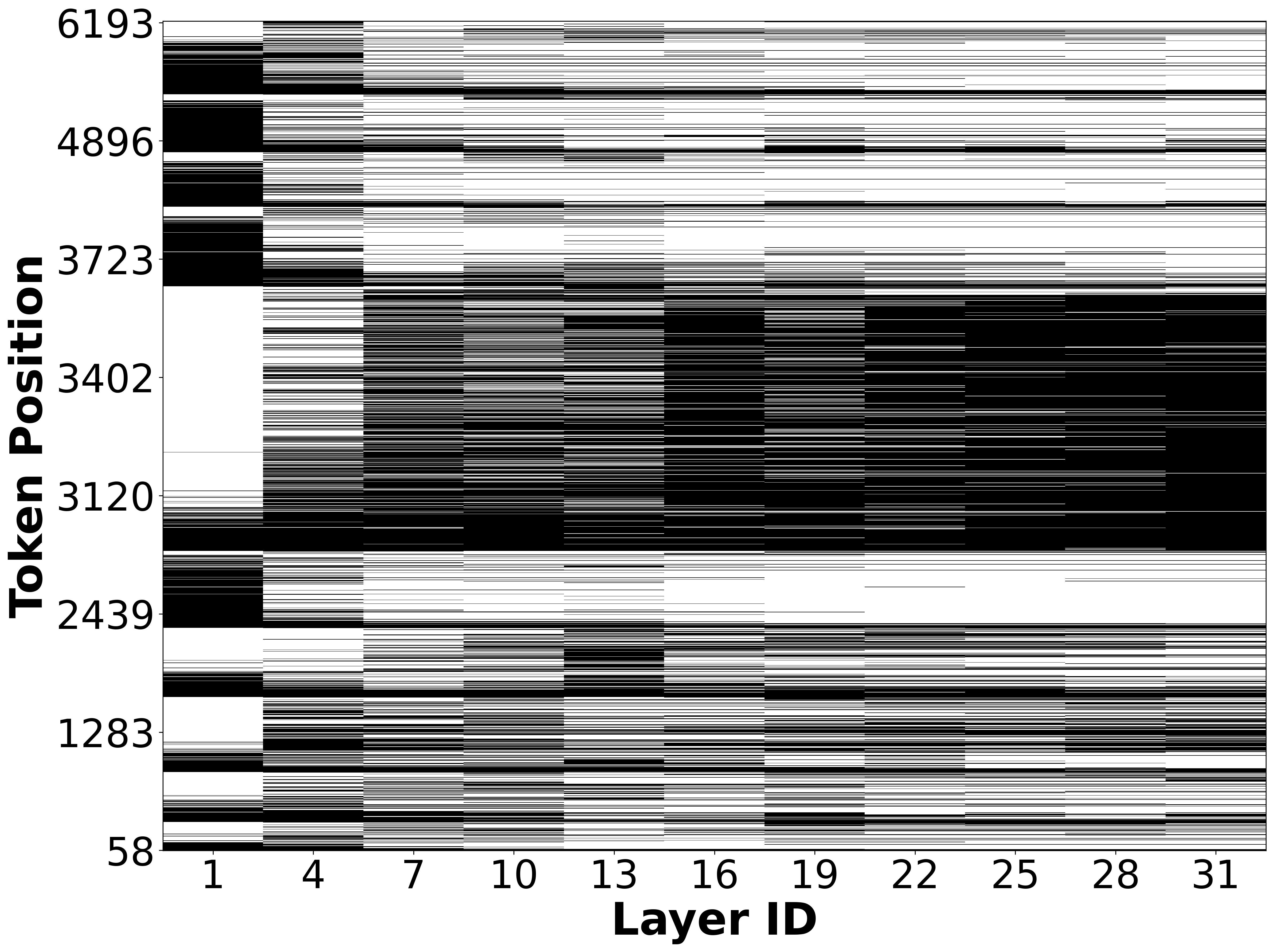}
  \end{minipage}
  \hfill
  \begin{minipage}[b]{0.48\linewidth}
    \centering
    \includegraphics[width=\linewidth]{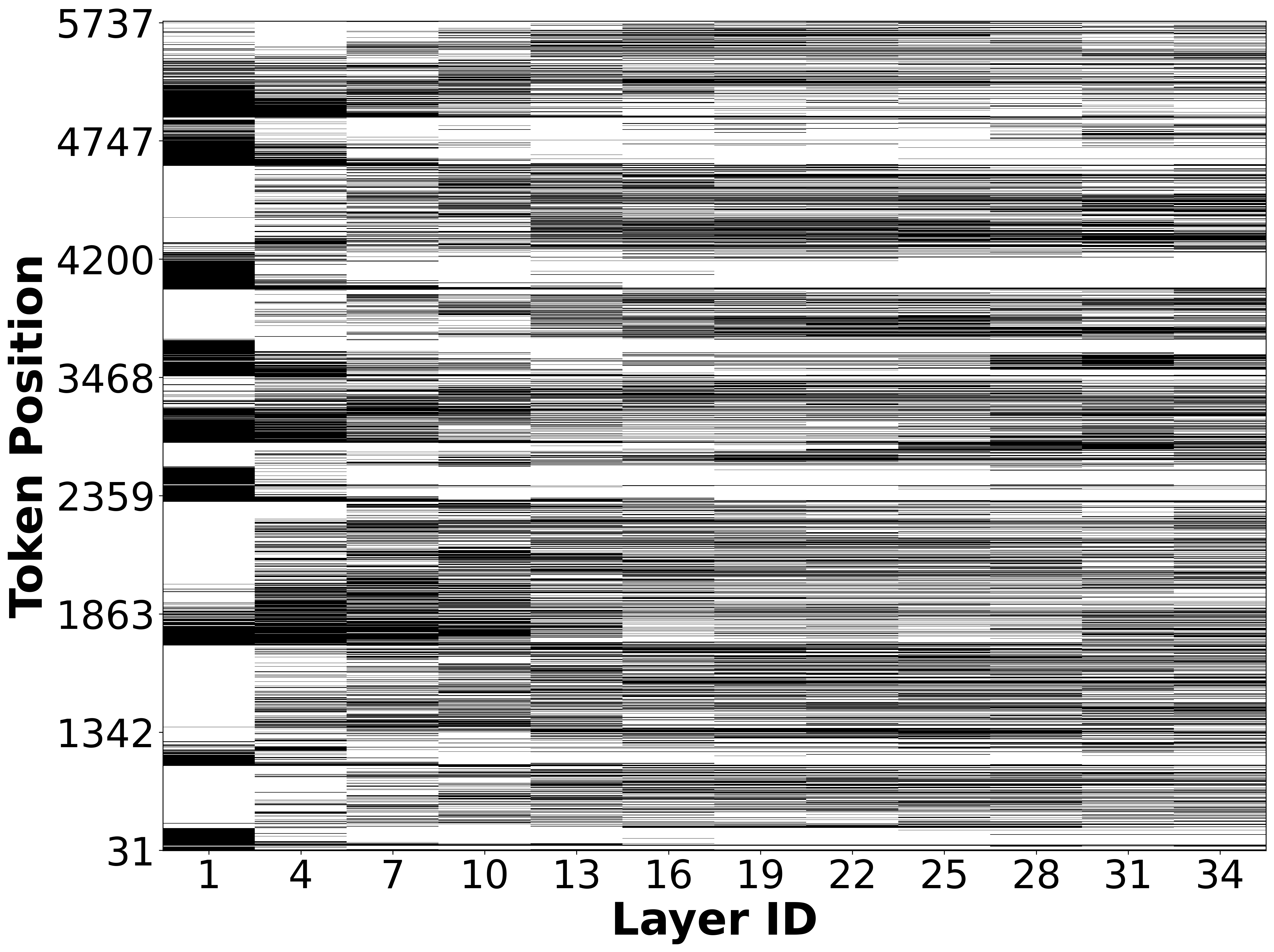}
  \end{minipage}
  \caption{Ideal token selection using $\Delta K$ ($r=15\%$) for one query in Llama-2WikiMQA (left) and  Qwen-Musique (right). Black/white: a token is/is not selected at that layer. The clusters of chosen tokens at early layers correspond to beginning-of-chunk tokens.}\label{fig:hm}
\end{figure}

To identify the causes of Cacheblend's accuracy loss with respect to full-prefill, we measure for each layer $L$ which are the $r\%$ of the tokens that exhibit the highest  $\Delta K$ at that layer, assuming a full prefill up to layer $L$. Figure~\ref{fig:hm} reports the tokens chosen for one random query from  2WikiMQA using Llama, and one random query from Musique  using Qwen. For clarity, the figure only reports the ``presence'' only of tokens that is in the top $r\%$ in at least one layer. 
We identify  two main observations that explain Cacheblend's accuracy and that, as we show later, are closely related with the design of following CLC systems.

\observation{Beginning of chunks tokens are critical} they exhibit high $\Delta K$ because of two phenomena: $i)$ initial prompt tokens tend to have very high attention scores, a phenomenon known as {\em attention sink}~\cite{xiao2024efficientstreaminglanguagemodels}. When computed separately, each chunk has such an attention sink; conversely, if computing a single prompt with all concatenated chunks, only one attention sink is present. $ii)$ Tokens are prone to attend more strongly to the closely preceding tokens (both structurally and as an artifact of positional encoding such as RoPE); tokens that are at the beginning of a chunk have no (or few) preceding tokens to attend to if the chunks are processed individually, differently from what happens if all the chunks are concatenated and processed as one context. 
\observation{Optimal token selection is dynamic} the tokens with highest $\Delta K$ at the first layer are not indicative of the tokens with high $\Delta K$ at following layers. For example, in our experiments only 8\% to 20\% of the tokens selected for recomputation are constant across all layers (for a given query), and they always appear at the beginning of a chunk. In addition, the tokens with high $\Delta K$ exhibit high variability also in deeper layers, as visible in Figure~\ref{fig:hm}.

\mypar{EPIC and Link0~\cite{hu2025epicefficientpositionindependentcaching}} EPIC  suggests that it is sufficient to recompute the KV caches of the first few tokens in each chunk to counter the attention sink effect and recover most of the full-prefill cross-chunk attention. 
Link0 is a lightweight variant of EPIC that  prepends $P$ prefix tokens (which can even be blank spaces) to each chunk when computing its KV cache, and then discards them when re-using the KV cache to serve a query. These $P$ tokens absorb the high attention  on the initial tokens in the chunk. Link0 does not recover any cross-chunk attention, but we discuss it now because its approach is adopted as a building block in other systems that we discuss later, and because it has been proposed jointly with EPIC. 

\begin{figure*}[t]
  \centering
  \includegraphics[height=3.8cm]{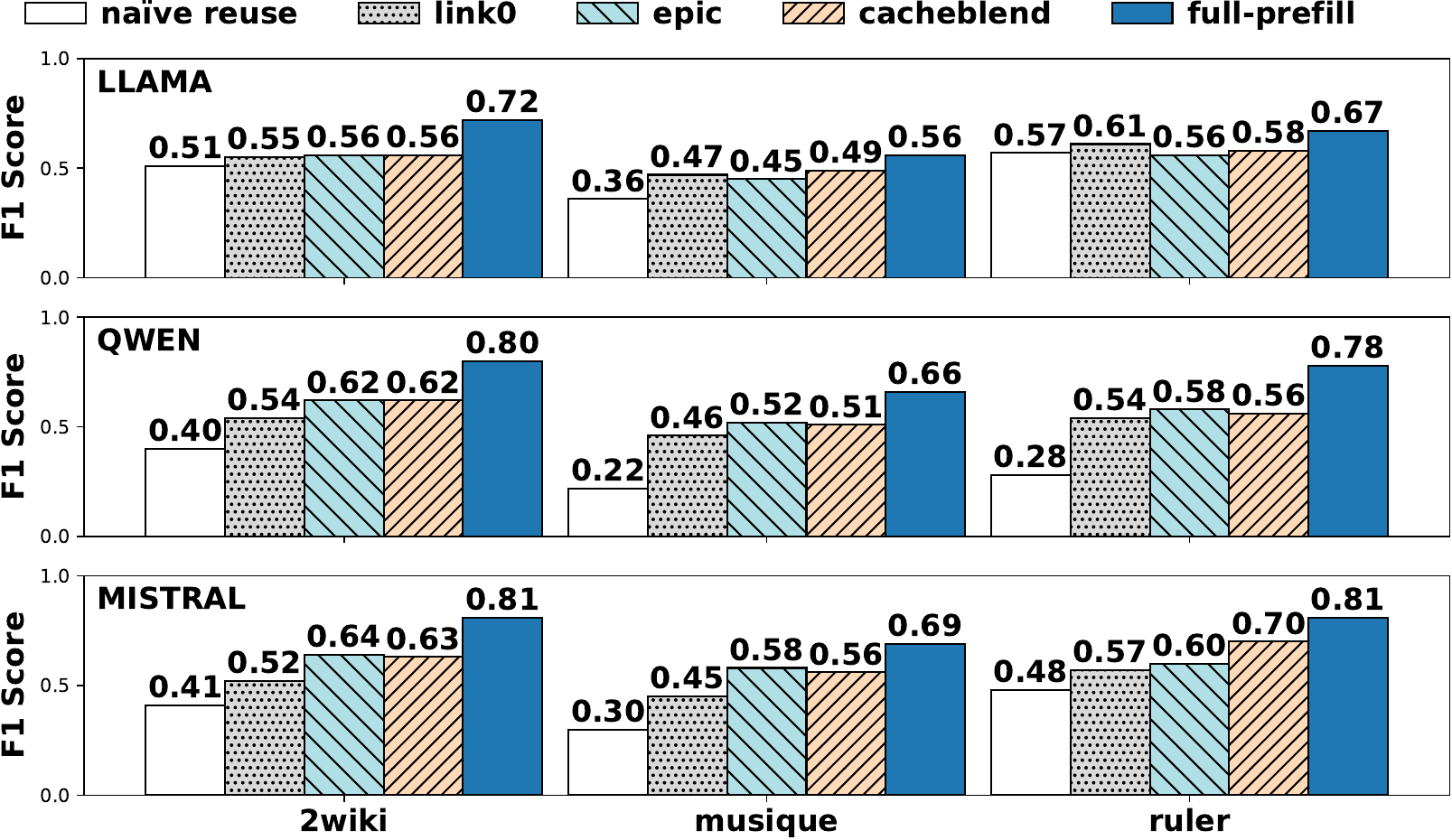}%
  \hfill
  \includegraphics[height=3.8cm]{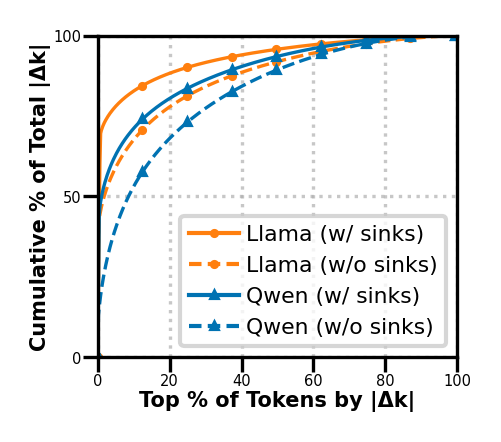}%
  \hfill
  \includegraphics[height=3.8cm]{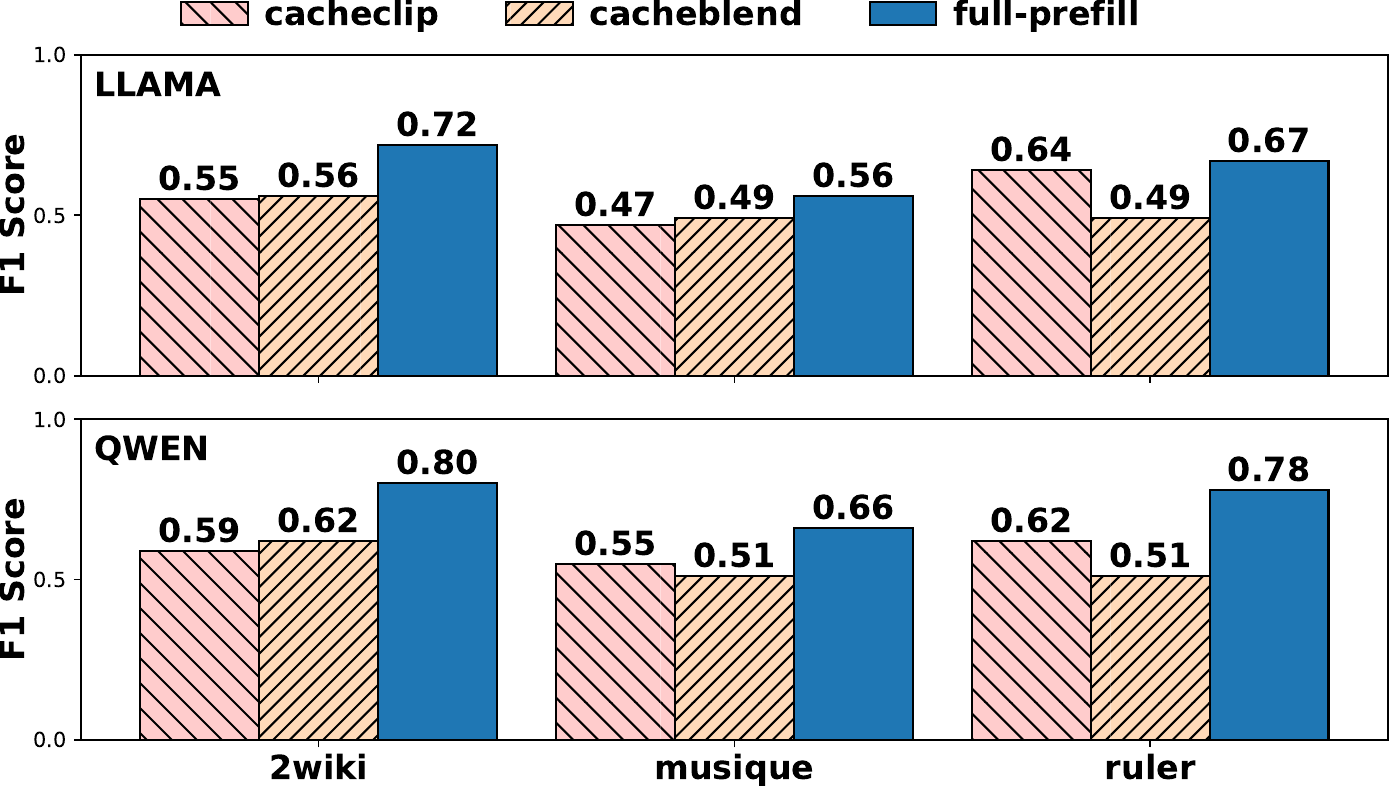}%
  \caption{%
    Left: EPIC/Link0 accuracy. %
    Center: cumulative percentage of tokens' $\Delta K$  for Llama on 2WikiMQA and Qwen on Musique, with and without the first 10 tokens (sink) of each chunk. %
    Right: CacheClip accuracy.%
  }
  \label{fig:combined}
\end{figure*}

Figure~\ref{fig:combined} (left) compares EPIC and Link0 to full prefill, naive reuse, and Cacheblend. We set $r=15\%$ for both EPIC and Cacheblend. For Link0 we perform tuning of $p$ in the set \{1,2,3,5,10\}, as suggested by recent work~\cite{yang2025apefasterlongercontextaugmented}. 
EPIC's accuracy is mostly on-par or even slightly better than Cacheblend, with only one case (Llama-Musique) where CacheBlend clearly outperforms EPIC. These results might come as a surprise, because EPIC restricts the set KV entries to be recomputed to the beginning of each chunks, while Cacheblend allows  more sparse recomputation. We argue that these results are in-line with our previous analysis of  Figure~\ref{fig:hm}: the $\Delta K$ at layer one, i.e., where Cacheblend performs its token choice, is heavily clustered around beginning-of-chunk tokens because of the attention sink and the locality of attention. Hence, Cacheblend and EPIC exhibit a high overlap in tokens being chosen for recomputation. 

Link0 outperforms naive-reuse, i.e., at the same zero recomputation overhead.  We provide low-level insights on the effectiveness of the attention sink removal of Link0 by reporting, in Figure~\ref{fig:combined} (middle), the cumulative $\Delta K$ accrued by context tokens, sorted by highest $Delta K$ value, with plain K cache reuse and in Link0. The plot reports results for LLama-2WikiMQA and Qwen-Musique, but similar results hold for other model/dataset combinations. The plot shows that artificially removing the sink tokens in Link0 helps in smoothing the $\Delta K$ distribution, lowering the initial spikes that correspond to the high $\Delta K$ concentration in sink tokens.  Despite this, Link0 is generally less accurate than EPIC and Cacheblend, except for Llama-Ruler, due to its lack of cross-chunk attention recovery. 

\observation{Addressing the attention sink is not enough} Absorbing the beginning-of-chunk attention leads to  attention states that follow a distribution more similar to the one of full-prefill. However, the  lack of any cross-attention recomputation ultimately limits the accuracy that this approach alone is able to deliver. Recovering the cross-attention only of beginning-of-chunk tokens is, similarly, not enough to achieve high accuracy. The heatmap in Figure~\ref{fig:hm} shows that, especially for mid and late layers, tokens with high $\Delta K$ do not appear only at the beginning of each chunk. %Hence, only recomputing the cross-attention only for tokens at the beginning of each chunk is not sufficient to recover full accuracy. 

\mypar{Cacheclip~\cite{yang2025cacheclipacceleratingrageffective} } Cacheclip is based on the hypothesis that deeper layers capture more meaningful attention dynamics, and therefore they are better suited to identify the tokens to recover cross-chunk attention. Cacheclip uses a small auxiliary model to serve the target query with the sole purpose of selecting the tokens for KV cache recomputation, using query attention scores at the last level as token importance metric. %To keep the additional latency of invoking such model low, Cacheclip uses a small auxiliary model, and computes the attention of the query to each chunk in parallel (i.e., without cross-chunk attention).  
 Cacheclip also adopts the same strategy as Link0 to address the attention sink phenomenon, i.e., by prepending $p$ prefix tokens when generating the chunk KV caches.
 
We run Cacheclip on Llama and Qwen using their small distilled versions (1B and 0.6B respectively) as auxiliary models to ensure tokenizer compatibility. We set  $r=15\%$ and we  tune the $p$ parameter in the set $\{1,2,3,5,10\}$.  Figure~\ref{fig:combined} (right) compares the accuracy of Cacheclip, Cacheblend and full-prefill. Cacheclip is slightly less accurate than Cacheblend ($\leq3\%$) in half of the cases, and clearly more accurate ($4\%$ to $15\%$) in the other half, at the cost of running an extra model to perform token selection. Cacheclip is especially effective on Ruler. We argue this is because Ruler adds chunks in the context that are irrelevant to the query, and Cacheclip attention-based selection approach is able to filter them out. Cacheclip, however, is in general not competitive with full-prefill.

\begin{figure}[b]
  \centering
  \includegraphics[width=\linewidth]{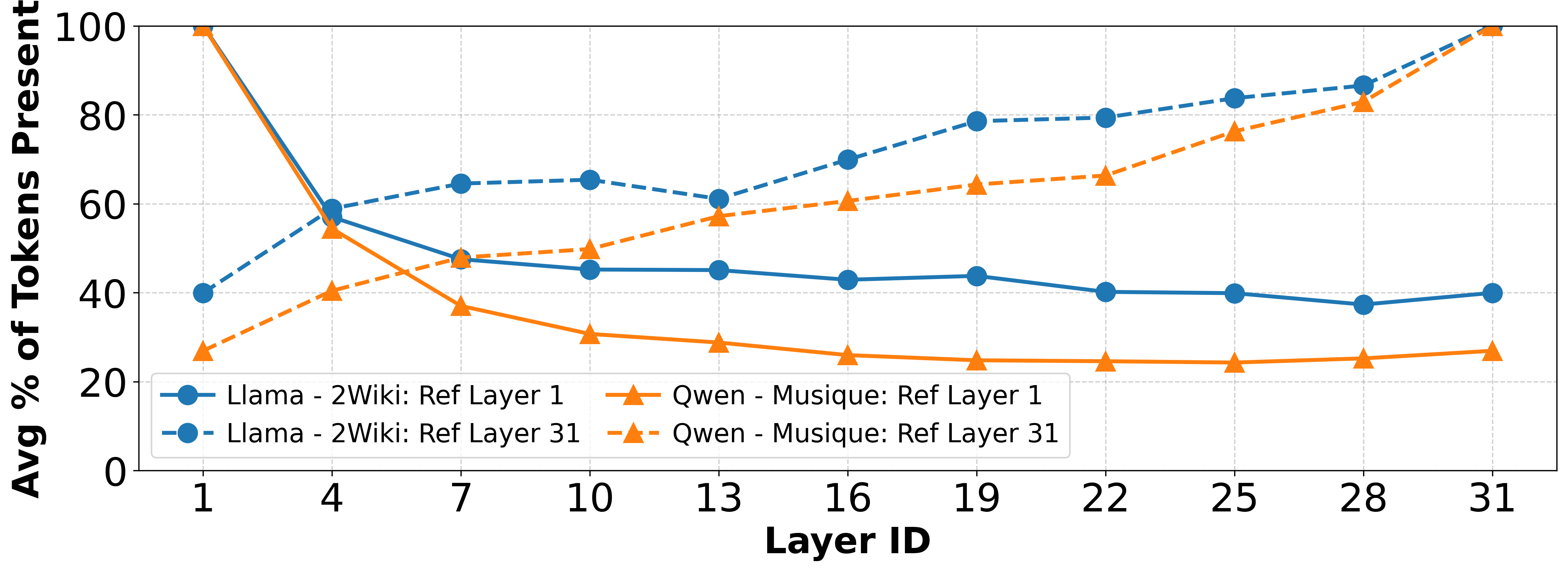}
  \caption{Average percentage of the tokens chosen (by $\Delta K$) for recomputation at a reference layer that would be chosen also in other layers ($r=15\%$).}
  \label{fig:ovlp}
\end{figure}

We investigate why Cacheclip is not always more accurate than Cacheblend, and why it is  worse than full-prefill on Musique and 2WikiMQA. To this end, we measure the quality of the token selection at the first layer and at a deeper layer (for visualization clarity we use the 31st layer as deep layer, which is the last for Llama and sixth-to-last in Qwen). We use the $\Delta K$ as metric for token selection, because it directly captures how the recomputed K caches differ from the full-prefill ones. Figure~\ref{fig:ovlp} reports the \% of tokens chosen at the first, resp. deeper, layer  that are also chosen in following, resp. preceding, layers, on Llama-2WikiMQA  and Qwen-Musique  using  $r=15\%$. On the one hand, tokens selected at the later layers overlap more with tokens selected at earlier layers than what tokens selected at the first layer do with later layers. This corroborates the insight of Cacheclip. On the other hand, however,  few tokens chosen at the deeper layer are among the tokens chosen at the early  layers. A poor token selection at the first layers might affect  the quality of the KV caches in the very  early layers of the LLM, which are crucial to pass high-quality activations to following layers. We argue that this is a key dynamic that limits Cacheclip's accuracy.

\observation{Last-layer token selection is not a silver bullet}  Cacheclip's token selection strategy  is more robust than Cacheblend's at deeper layers, but  not at early ones, which limits its overall accuracy. Moreover, integrating a second model in an LLM inference system has performance and architectural implications that we do not address in this study, but that warrant investigation. Overall, we argue that last-layer token selection can be a building block of a dynamic token selection scheme, coupled with first-layer token selection for high KV cache quality in early layers.

\mypar{Droidspeak~\cite{liu2025droidspeakkvcachesharing}} Droidspeak  is a system proposed for KV cache re-use across fine-tuned models that share the same original model. The main insight of Droidspeak is that a model (the receiver) can serve a query by re-using most of the KV caches produced with a  full prefill by another model (the donor) to answer the {\em same} query. Droidspeak identifies two {\em critical layers}, $L_s$ and $L_e$, for a receiver model: when serving a query, KV caches from the donor model up to $L_s$ are re-used; then KV caches from $L_s$ to $L_e$ are recomputed as in full prefill; finally,  from $L_e$ onward KV caches from the donor model are re-used. To be able to switch from full re-use to full recomputation at layer $L_s$, Droidspeak caches the activations of the chunks at layer $L_e$ in the donor model, which are used by the receiving model to compute the $Q, K, V$ matrices.

Droidspeak can be adapted to implement a dynamic choice of the tokens for selective KV cache recomputation. Such approach would overcome the limitations of the static choice which hinder the accuracy of the approaches seen so far. We test Droidspeak's approach adapting it to our target CLC scenario. First, we do not assume repeating queries across similar models. Crucially, this means that the activations cached and reused  are obtained from the individual prefills of the KV caches, which do not have any cross-chunk attention. Second, we augment Droidspeak with KV cache selective recomputation, instead of the full-reuse/full-recompute paradigm. In our experiments,  Droidspeak recomputes $15\%$ of the KV cache entries at layer one, and then it re-selects the tokens for recomputation at layer 5. In both cases, Droidspeak uses $\Delta K$ to rank tokens for recomputation.  We choose layer 5 because we have observed that at that layer the $\Delta K$ scores are not too concentrated anymore around beginning-of-chunk tokens for the cases we consider. This allows the recomputation mechanism to choose a set of tokens that is more relevant to recover cross-chunk attention in deeper layers.
\begin{figure}[t]
    \includegraphics[width=\linewidth]{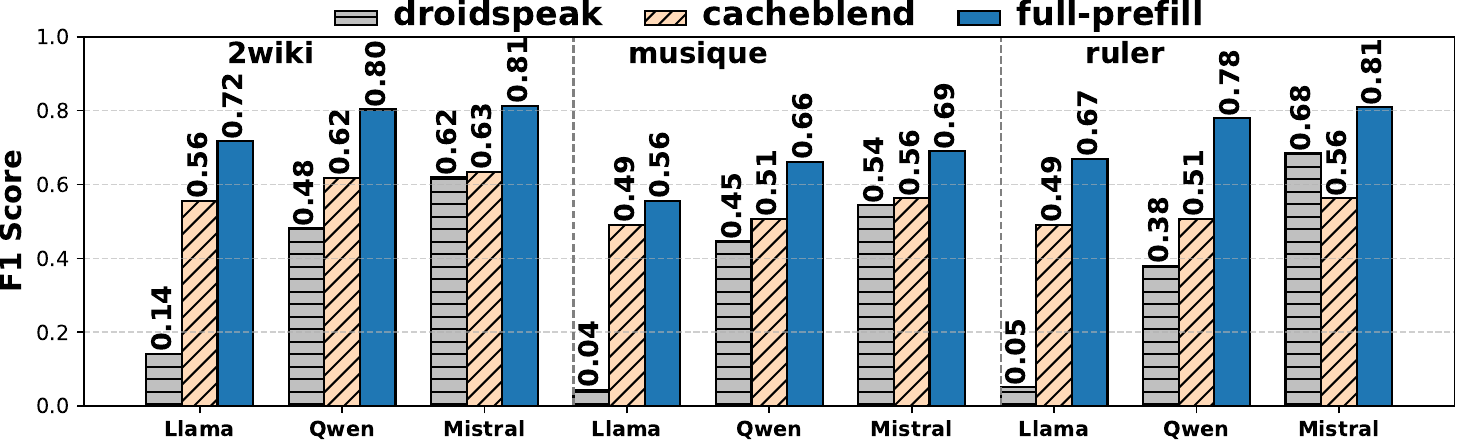}
  \caption{Droidspeak accuracy.}\label{fig:droidspeak}
\end{figure}

Figure~\ref{fig:droidspeak} reports the results we obtained with Droidspeak, using Cacheblend (with $r=15\%$) and full-prefill as comparison. With the exception of Qwen-Musique, Droidspeak achieves worse accuracy than Cacheblend. Droidspeak achieves very low accuracy on Llama, where it leads the model to output junk tokens in many queries. 
\observation{Dynamic recomputation needs cached cross-attention} we ascribe the poor accuracy of Droidspeak design in our CLC scenario to the lack of cross-chunk attention in the cached activations. Droidspeak achieves good accuracy in the case where activations can be cached with full cross-attention, albeit from a different donor model~\cite{liu2025droidspeakkvcachesharing}. An interesting research avenue is to evaluate dynamic recomputation in designs that preserve cross-attention in the cached KV caches, which we discuss in Section~\ref{sec:conclusion}.

\takeawaybox{1}{Recomputation-based CLC approaches are not competitive with full-prefill. Static strategies identify sub-optimal KV cache entries to recompute; dynamic ones lack cross-chunk attention information in cached attention states.}

\subsection{Attention-reshaping mechanism}
In this section we discuss approaches that aim to improve CLC systems accuracy by obtaining attention dynamics similar to full-prefill without relying on cross-chunk attention.

\mypar{APE~\cite{yang2025apefasterlongercontextaugmented} and SEL~\cite{zhang2025attentionentropykeyfactor}} These two systems make the observation that the attention scores in CLC systems are different from the ones in full prefill for two reasons. First,  attention sinks appear in every individually computed KV cache, and only once in the KV caches in full-prefill. Second, the query tends to attend more uniformly to tokens in each chunk in the CLC case than in  the full-prefill case, where the query attention is mostly concentrated on closer context tokens and sparsely on more distant context tokens. 
To fix the first issue, both systems adopt the same approach as Link0 of prepending additional prefix tokens. For the second issue, APE computes the attention of the query towards context tokens using a scaling factor (lower temperature, $t$) in the softmax operator, and then multiplies the attention scores by another scaling factor ($s$). SEL, instead, lets the query attend only to a subset of the chunks, with the goal or reducing attention entropy, which is a factor that negatively correlates with accuracy.  We choose APE as representative system of this CLC approach, since its code is available and it is easy to integrate in our framework. We perform a simple hyper-parameter tuning approach to find good values for the number of additional prefix tokens, temperature, and scaling factor to use. 
\begin{figure}[t]
    \includegraphics[width=\linewidth]{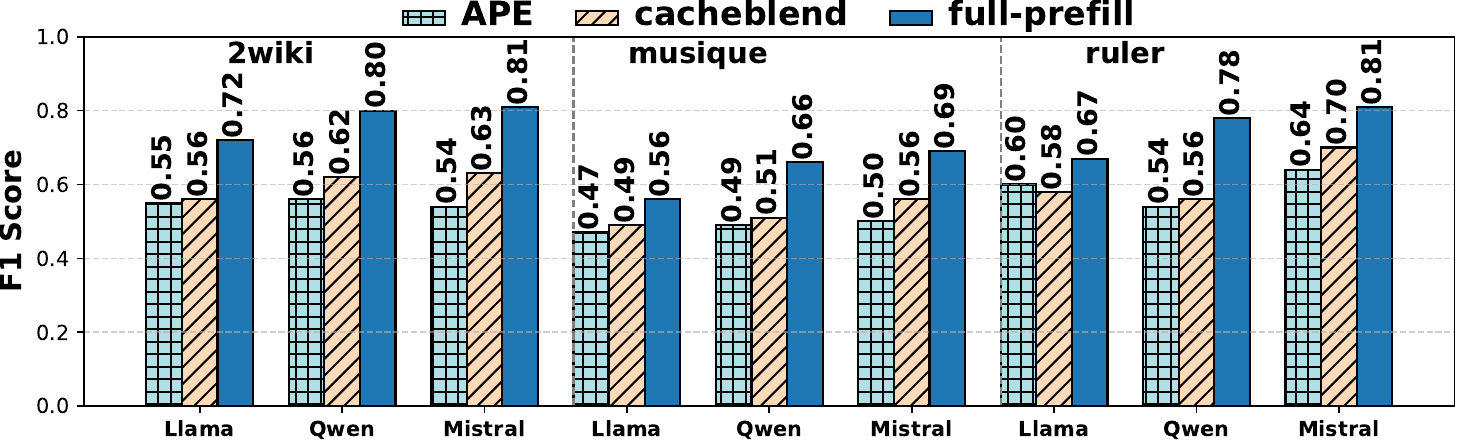}
  \caption{APE accuracy.}\label{fig:ape}
\end{figure}

We tune APE by choosing the $t$, $s$, and $p$ parameters with a simple grid search as in the original paper.
Figure~\ref{fig:ape} compares the accuracy of APE against full-prefill and Cacheblend. APE's accuracy  is 2\% to 9\% lower than Cacheblend's in all cases except in Llama-Ruler, where it is better by just 1\%.  It is important to note that APE does not require any KV cache entry recomputation, and its attention transformation can be integrated with efficient attention computation, which results in a lower computational overhead than Cacheblend. With respect to naive-reuse, which also has no recomputation overhead, APE is consistently better. 
\observation{Cross-chunk attention is paramount} approaches that improve the query-to-cached-chunks attention eschew the recomputation overhead, but the lack of cross-attention ultimately cripples their accuracy.

\mypar{TurboRAG~\cite{lu2024turboragacceleratingretrievalaugmentedgeneration}, BlockAttention~\cite{ma2025blockattentionefficientprefilling}, KVLink~\cite{yang2025kvlinkacceleratinglargelanguage}} These systems  fine-tune the  LLM to train the attention layers to not rely on cross-chunk attention. This is accomplished by using attention matrices where only intra-chunks attention is allowed for context tokens. KVLink, additionally, inserts, at fine-tuning and at query time,  a few {\em link tokens} with full attention scope before each context, with the goal of recovering partially the cross-chunk attention.

\begin{figure}[t]
  \centering
    \includegraphics[width=\linewidth]{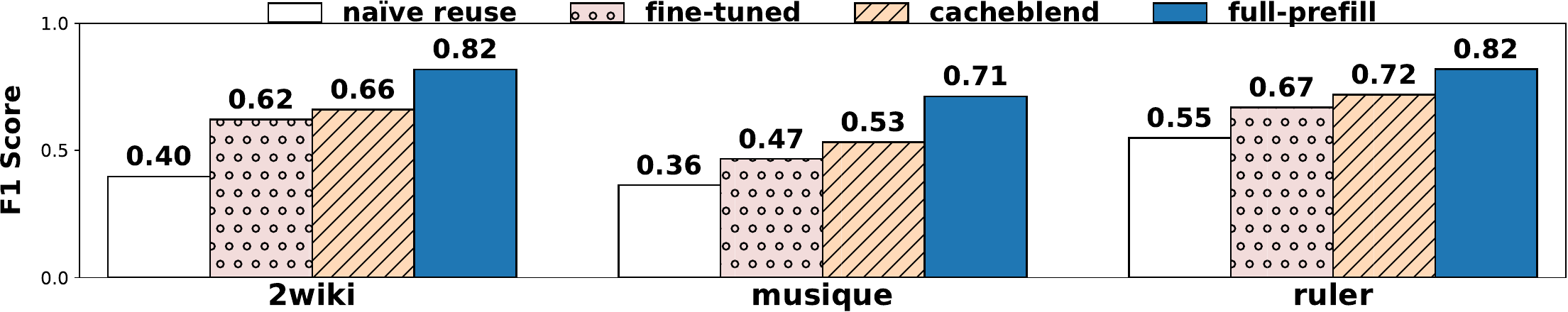}
  \caption{BlockAttention's accuracy on fine-tuned LLama.}\label{fig:ft}
\end{figure}

The fine-tuning processes of these systems require a significant amount of data and compute (e.g., 30 A100 GPUs for TurboRAG and more than 10 million data points for KVLink), which makes replicating their approaches cumbersome.  To evaluate the fine-tuning approach on our testbed we first implement the Turborag/BlockAttention approach: we fine-tune our target models using a LoRA-based~\cite{hu2021loralowrankadaptationlarge} approach using 2000 samples from a Q\&A dataset taken from an extended, disjoint 2WikimMQA dataset. The resulting models, however, gives accuracy that is comparable to that of naive KV cache reuse. Then, we use the BlockAttention models available on HuggingFace, based on Llama3.1-8B, and we evaluate them on our test datasets.  Figure~\ref{fig:ft} reports the accuracy achieved by the baseline model, Cacheblend with $r=15\%$ on the baseline model, naive reuse and the fine-tuned BlockAttention model. The fine-tuned model recovers much accuracy with respect to the naive-reuse case, at the same null recomputation cost, but achieves 4 to 7\% lower accuracy than Cacheblend, which however incurs recomputation overhead.

\observation{Fine tuning approaches are challenging to implement} they require substantial effort, computing power and data to adapt to the target workload. % query distribution and attention patterns. 
The solutions we tried are not competitive with recomputation-based approaches, arguably because of the lack of cross-attention recomputation.

\takeawaybox{2}{The accuracy of attention-reshaping techniques is not competitive with recomputation-based ones. The lack of cross-chunk attention recomputation is the limiting factor.}

\section{Exploring synergies among designs}\label{sec:golem}
We show that existing design CLC choices, proposed independently, are in fact synergistic, and that it is possible to craft a system that, by combining them, can achieve higher accuracy. We note in particular that $i)$ recovering cross-chunk attention always improves accuracy; $ii)$ addressing the attention sink phenomenon is paramount to avoid artificially skewed $\Delta K$ scores; and $iii)$ scaling the context attention temperature as done in APE boosts accuracy significantly. We combine these elements in a design that we refer to as {\em PSR} (Prefix-Scale-Recompute). PSR removes the sink tokens by appending $p$ blank prefix tokens as in Link0, selects the $r\%$ tokens to recompute as in Cacheblend, and scales the attention to the context tokens as in APE. PSR does not scale the attention scores for the recomputed tokens, because it recovers their cross-chunk attention, and then treats them like APE treats query tokens.
We test PSR using $r=15\%$ and performing similar hyper parameter tuning for the prefix, temperature and scaling parameters as discussed in previous sections. We also perform and ablation study of the contribute of each design component to accuracy, including R (equivalent to Cacheblend), S (equivalent to Link0), S+P (as in Link0+APE), and P+S+R (our new design).

\begin{figure}[t]
\centering
    \includegraphics[width=1\linewidth]{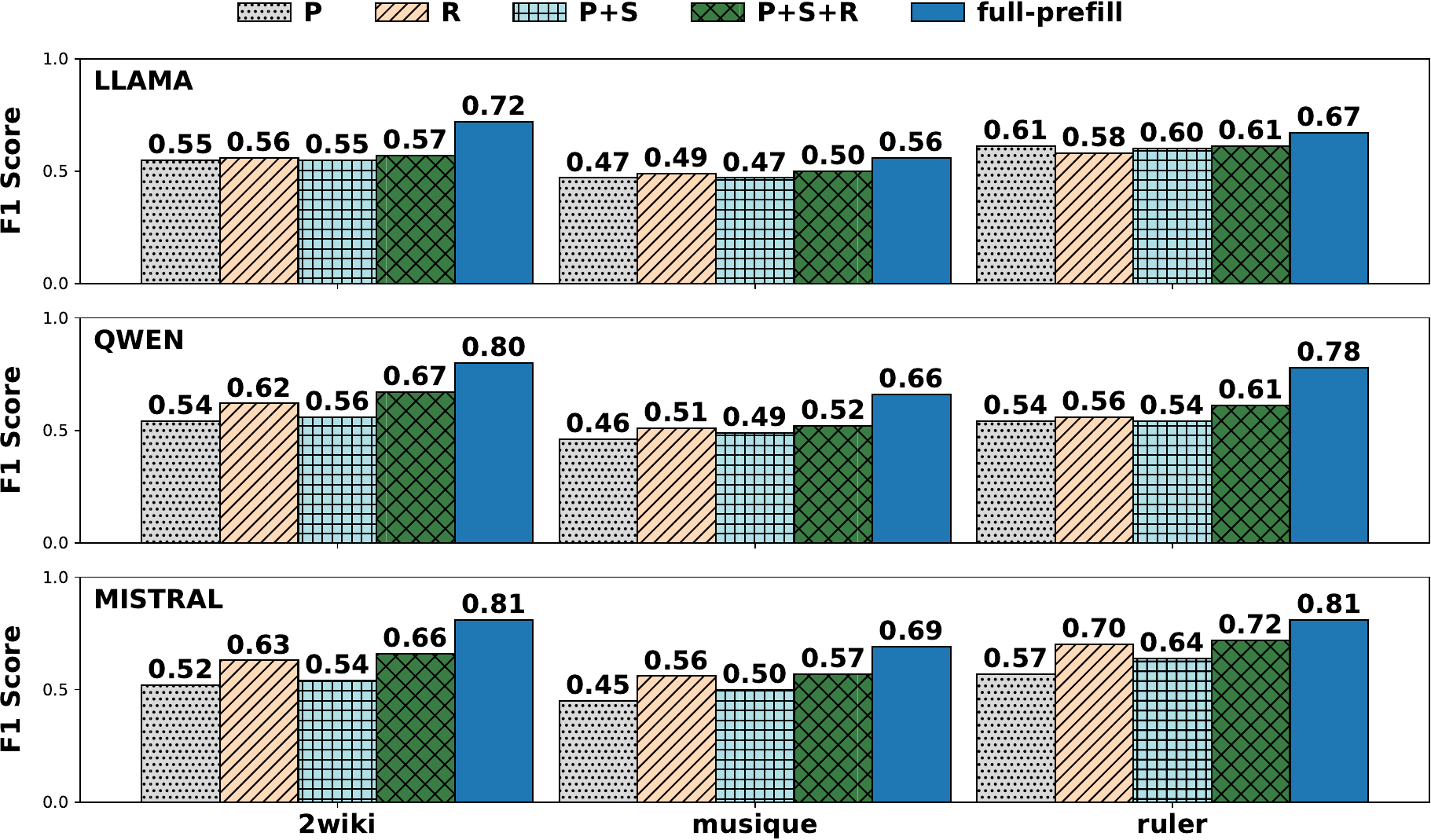}
  \caption{PSR: accuracy and ablation study.}\label{fig:psr}
\end{figure}

Figure~\ref{fig:psr} reports the results of our experiments. P+S+R is consistently the best performing design, with  improvements up to 5\% in absolute accuracy score. Our ablation study also confirms that the combination of all techniques is instrumental to achieving high accuracy in a robust fashion: neither P+S nor R alone are always the best second choice, indicating that relying on either of them may achieve sub-optimal accuracy in some workloads.

Our current PSR design does not incorporate dynamic token selection or deeper layer token selection. Our key observations, discussed in detail in Section 3, reveal significant technical challenges that currently prevent the effective integration of these advanced features. We believe that overcoming these challenges is crucial and will be the key to unlocking higher accuracy through dynamic, multi-layer token selection in future work.

\vspace{-5pt}
\takeawaybox{3}{Attention-reshaping mechanisms + selective KV cache recomputation achieves higher accuracy than the two techniques singularly. Enabling dynamic token selection can potentially further narrow the gap to full-prefill accuracy.}

\section{Conclusion}\label{sec:conclusion}
We discussed and compared CLC approaches based on cross-chunk attention recovery  and attention reshaping approaches. We showed that recovering cross-chunk  attention is paramount for accuracy. We also showed that existing selective recomputation techniques aimed to recover cross-attention have fundamental limitations: static approaches identify suboptimal KV cache entries to recompute; dynamic ones cannot incorporate cross-chunk attention effectively. We finally showed that selective KV cache recomputation approach and attention reshaping exhibit complementary strengths.  We designed a system that combines the best of both worlds and achieves up to 5\% better accuracy.

 \mypar{Future directions}  We performed our study  on single-shot Q\&A tasks based on human queries. As the applicability of LLMs spreads to different domains, it is interesting to evaluate the accuracy achieved by CLC techniques on other tasks, e.g., coding and summarization, multi-turn chatbots, and in agentic workloads.
 
 We focused on the most prominent approach to CLC, which entails computing the KV caches of text chunks individually. A new design, spearheaded by the Cachecraft system ~\cite{agarwal2025cachecraftmanagingchunkcachesefficient}, departs from this approach, and  obtains the KV caches from full-prefill runs on previous queries.  Cachecraft obtains KV caches that store cross-chunk attention, which we have identified as a key requirement to unlock high accuracy.  On the flip side, Cachecraft stores different KV caches of the same chunk, each corresponding to a different contextualization, to be able to choose the most compatible version with the target query.
 Investigating, evaluating and improving  Cachecraft's design  requires the definition of more complex CLC benchmarks where the same chunks are reused multiple times, with different contexts, and with different reuse patterns. Such benchmarks would enable the analysis of the storage vs accuracy trade-off incurred by storing multiple KV caches for the same chunk, and the exploration of effective cache selection heuristics.
 
We identify as promising research areas the design of systems that leverage cached cross-chunk attention and dynamic token selection for recomputation, and new benchmarking tools to evaluate different kind of workloads and KV cache re-use patterns.

\bibliographystyle{unsrt}
\bibliography{dac}

@inproceedings{yao2025cacheblendfastlargelanguage,
author = {Yao, Jiayi and Li, Hanchen and Liu, Yuhan and Ray, Siddhant and Cheng, Yihua and Zhang, Qizheng and Du, Kuntai and Lu, Shan and Jiang, Junchen},
title = {CacheBlend: Fast Large Language Model Serving for RAG with Cached Knowledge Fusion},
year = {2025},
isbn = {9798400711961},
publisher = {Association for Computing Machinery},
address = {New York, NY, USA},
url = {https://doi.org/10.1145/3689031.3696098},
doi = {10.1145/3689031.3696098},
booktitle = {Proceedings of the Twentieth European Conference on Computer Systems},
pages = {94–109},
numpages = {16},
location = {Rotterdam, Netherlands},
series = {EuroSys '25}
}

@article{yang2025cacheclipacceleratingrageffective,
  author       = {Bin Yang and
                  Qiuyu Leng and
                  Jun Zeng and
                  Zhenhua Wu},
  title        = {CacheClip: Accelerating {RAG} with Effective {KV} Cache Reuse},
  journal      = {CoRR},
  volume       = {abs/2510.10129},
  year         = {2025},
  url          = {https://doi.org/10.48550/arXiv.2510.10129},
  doi          = {10.48550/ARXIV.2510.10129},
  eprinttype    = {arXiv},
  eprint       = {2510.10129},
  bibsource    = {dblp computer science bibliography, https://dblp.org}
}

@article{agarwal2025cachecraftmanagingchunkcachesefficient,
      author       = {Shubham Agarwal and
                  Sai Sundaresan and
                  Subrata Mitra and
                  Debabrata Mahapatra and
                  Archit Gupta and
                  Rounak Sharma and
                  Nirmal Joshua Kapu and
                  Tong Yu and
                  Shiv Kumar Saini},
  title        = {Cache-Craft: Managing Chunk-Caches for Efficient Retrieval-Augmented
                  Generation},
  journal      = {Proc. {ACM} Manag. Data},
  volume       = {3},
  number       = {3},
  pages        = {136:1--136:28},
  year         = {2025},
  url          = {https://doi.org/10.1145/3725273},
  doi          = {10.1145/3725273},
  bibsource    = {dblp computer science bibliography, https://dblp.org}
}

@inproceedings{yang2025apefasterlongercontextaugmented,
  author       = {Xinyu Yang and
                  Tianqi Chen and
                  Beidi Chen},
  title        = {{APE:} Faster and Longer Context-Augmented Generation via Adaptive
                  Parallel Encoding},
  booktitle    = {The Thirteenth International Conference on Learning Representations,
                  {ICLR} 2025, Singapore, April 24-28, 2025},
  publisher    = {OpenReview.net},
  year         = {2025},
  url          = {https://openreview.net/forum?id=yUC8pU508S},
  bibsource    = {dblp computer science bibliography, https://dblp.org}
}

@article{hu2025epicefficientpositionindependentcaching,
      author       = {Junhao Hu and
                  Wenrui Huang and
                  Haoyi Wang and
                  Weidong Wang and
                  Tiancheng Hu and
                  Qin Zhang and
                  Hao Feng and
                  Xusheng Chen and
                  Yizhou Shan and
                  Tao Xie},
  title        = {{EPIC:} Efficient Position-Independent Context Caching for Serving
                  Large Language Models},
  journal      = {CoRR},
  volume       = {abs/2410.15332},
  year         = {2024},
  url          = {https://doi.org/10.48550/arXiv.2410.15332},
  doi          = {10.48550/ARXIV.2410.15332},
  eprinttype    = {arXiv},
  eprint       = {2410.15332},
  bibsource    = {dblp computer science bibliography, https://dblp.org}
}

@inproceedings{vaswani2023attentionneed,
      author       = {Ashish Vaswani and
                  Noam Shazeer and
                  Niki Parmar and
                  Jakob Uszkoreit and
                  Llion Jones and
                  Aidan N. Gomez and
                  Lukasz Kaiser and
                  Illia Polosukhin},
  editor       = {Isabelle Guyon and
                  Ulrike von Luxburg and
                  Samy Bengio and
                  Hanna M. Wallach and
                  Rob Fergus and
                  S. V. N. Vishwanathan and
                  Roman Garnett},
  title        = {Attention is All you Need},
  booktitle    = {Advances in Neural Information Processing Systems 30: Annual Conference
                  on Neural Information Processing Systems 2017, December 4-9, 2017,
                  Long Beach, CA, {USA}},
  pages        = {5998--6008},
  year         = {2017},
  url          = {https://proceedings.neurips.cc/paper/2017/hash/3f5ee243547dee91fbd053c1c4a845aa-Abstract.html},
  bibsource    = {dblp computer science bibliography, https://dblp.org}
}

@article{liu2025droidspeakkvcachesharing,
      title={DroidSpeak: KV Cache Sharing for Cross-LLM Communication and Multi-LLM Serving}, 
      author={Yuhan Liu and Yuyang Huang and Jiayi Yao and Shaoting Feng and Zhuohan Gu and Kuntai Du and Hanchen Li and Yihua Cheng and Junchen Jiang and Shan Lu and Madan Musuvathi and Esha Choukse},
      year={2025},
      eprint={2411.02820},
      archivePrefix={arXiv},
      primaryClass={cs.MA},
      url={https://arxiv.org/abs/2411.02820}, 
}

@article{lu2024turboragacceleratingretrievalaugmentedgeneration,
      author       = {Songshuo Lu and
                  Hua Wang and
                  Yutian Rong and
                  Zhi Chen and
                  Yaohua Tang},
  title        = {TurboRAG: Accelerating Retrieval-Augmented Generation with Precomputed
                  {KV} Caches for Chunked Text},
  journal      = {CoRR},
  volume       = {abs/2410.07590},
  year         = {2024},
  url          = {https://doi.org/10.48550/arXiv.2410.07590},
  doi          = {10.48550/ARXIV.2410.07590},
  eprinttype    = {arXiv},
  eprint       = {2410.07590},
  bibsource    = {dblp computer science bibliography, https://dblp.org}
}

@inproceedings{ma2025blockattentionefficientprefilling,
      author       = {Dongyang Ma and
                  Yan Wang and
                  Tian Lan},
  title        = {Block-Attention for Efficient Prefilling},
  booktitle    = {The Thirteenth International Conference on Learning Representations,
                  {ICLR} 2025, Singapore, April 24-28, 2025},
  publisher    = {OpenReview.net},
  year         = {2025},
  url          = {https://openreview.net/forum?id=7zNYY1E2fq},
  bibsource    = {dblp computer science bibliography, https://dblp.org}
}

@article{yang2025kvlinkacceleratinglargelanguage,
      author       = {Jingbo Yang and
                  Bairu Hou and
                  Wei Wei and
                  Yujia Bao and
                  Shiyu Chang},
  title        = {KVLink: Accelerating Large Language Models via Efficient {KV} Cache
                  Reuse},
  journal      = {CoRR},
  volume       = {abs/2502.16002},
  year         = {2025},
  url          = {https://doi.org/10.48550/arXiv.2502.16002},
  doi          = {10.48550/ARXIV.2502.16002},
  eprinttype    = {arXiv},
  eprint       = {2502.16002},
  bibsource    = {dblp computer science bibliography, https://dblp.org} 
}

@inproceedings{lewis2021retrievalaugmentedgenerationknowledgeintensivenlp,
      author       = {Patrick Lewis and
                  Ethan Perez and
                  Aleksandra Piktus and
                  Fabio Petroni and
                  Vladimir Karpukhin and
                  Naman Goyal and
                  Heinrich K{\"{u}}ttler and
                  Mike Lewis and
                  Wen{-}tau Yih and
                  Tim Rockt{\"{a}}schel and
                  Sebastian Riedel and
                  Douwe Kiela},
  editor       = {Hugo Larochelle and
                  Marc'Aurelio Ranzato and
                  Raia Hadsell and
                  Maria{-}Florina Balcan and
                  Hsuan{-}Tien Lin},
  title        = {Retrieval-Augmented Generation for Knowledge-Intensive {NLP} Tasks},
  booktitle    = {Advances in Neural Information Processing Systems 33: Annual Conference
                  on Neural Information Processing Systems 2020, NeurIPS 2020, December
                  6-12, 2020, virtual},
  year         = {2020},
  url          = {https://proceedings.neurips.cc/paper/2020/hash/6b493230205f780e1bc26945df7481e5-Abstract.html},
  bibsource    = {dblp computer science bibliography, https://dblp.org}
}

@inproceedings{ho2020constructingmultihopqadataset,
      author       = {Xanh Ho and
                  Anh{-}Khoa Duong Nguyen and
                  Saku Sugawara and
                  Akiko Aizawa},
  editor       = {Donia Scott and
                  N{\'{u}}ria Bel and
                  Chengqing Zong},
  title        = {Constructing {A} Multi-hop {QA} Dataset for Comprehensive Evaluation
                  of Reasoning Steps},
  booktitle    = {Proceedings of the 28th International Conference on Computational
                  Linguistics, {COLING} 2020, Barcelona, Spain (Online), December 8-13,
                  2020},
  pages        = {6609--6625},
  publisher    = {International Committee on Computational Linguistics},
  year         = {2020},
  url          = {https://doi.org/10.18653/v1/2020.coling-main.580},
  doi          = {10.18653/V1/2020.COLING-MAIN.580},
  bibsource    = {dblp computer science bibliography, https://dblp.org}
}

@article{trivedi2022musiquemultihopquestionssinglehop,
      author       = {Harsh Trivedi and
                  Niranjan Balasubramanian and
                  Tushar Khot and
                  Ashish Sabharwal},
  title        = {MuSiQue: Multihop Questions via Single-hop Question
                  Composition},
  journal      = {Trans. Assoc. Comput. Linguistics},
  volume       = {10},
  pages        = {539--554},
  year         = {2022},
  url          = {https://doi.org/10.1162/tacl\_a\_00475},
  doi          = {10.1162/TACL\_A\_00475},
  bibsource    = {dblp computer science bibliography, https://dblp.org} 
}

@article{hsieh2024rulerwhatsrealcontext,
      author       = {Cheng{-}Ping Hsieh and
                  Simeng Sun and
                  Samuel Kriman and
                  Shantanu Acharya and
                  Dima Rekesh and
                  Fei Jia and
                  Yang Zhang and
                  Boris Ginsburg},
  title        = {{RULER:} What's the Real Context Size of Your Long-Context Language
                  Models?},
  journal      = {CoRR},
  volume       = {abs/2404.06654},
  year         = {2024},
  url          = {https://doi.org/10.48550/arXiv.2404.06654},
  doi          = {10.48550/ARXIV.2404.06654},
  eprinttype    = {arXiv},
  eprint       = {2404.06654},
  bibsource    = {dblp computer science bibliography, https://dblp.org}
}

@inproceedings{gim2024promptcachemodularattention,
      author       = {In Gim and
                  Guojun Chen and
                  Seung{-}Seob Lee and
                  Nikhil Sarda and
                  Anurag Khandelwal and
                  Lin Zhong},
  editor       = {Phillip B. Gibbons and
                  Gennady Pekhimenko and
                  Christopher De Sa},
  title        = {Prompt Cache: Modular Attention Reuse for Low-Latency Inference},
  booktitle    = {Proceedings of the Seventh Annual Conference on Machine Learning and
                  Systems, MLSys 2024, Santa Clara, CA, USA, May 13-16, 2024},
  publisher    = {mlsys.org},
  year         = {2024},
  url          = {https://proceedings.mlsys.org/paper\_files/paper/2024/hash/a66caa1703fe34705a4368c3014c1966-Abstract-Conference.html},
  bibsource    = {dblp computer science bibliography, https://dblp.org} 
}

@article{grattafiori2024llama3herdmodels,
      author       = {Llama Team},
  title        = {The Llama 3 Herd of Models},
  journal      = {CoRR},
  volume       = {abs/2407.21783},
  year         = {2024},
  url          = {https://doi.org/10.48550/arXiv.2407.21783},
  doi          = {10.48550/ARXIV.2407.21783},
  eprinttype    = {arXiv},
  eprint       = {2407.21783},
  bibsource    = {dblp computer science bibliography, https://dblp.org}
}

@article{yang2025qwen3technicalreport,
      author       = {An Yang and
                  Anfeng Li and
                  Baosong Yang and
                  Beichen Zhang and
                  Binyuan Hui and
                  Bo Zheng and
                  Bowen Yu and
                  Chang Gao and
                  Chengen Huang and
                  Chenxu Lv and
                  Chujie Zheng and
                  Dayiheng Liu and
                  Fan Zhou and
                  Fei Huang and
                  Feng Hu and
                  Hao Ge and
                  Haoran Wei and
                  Huan Lin and
                  Jialong Tang and
                  Jian Yang and
                  Jianhong Tu and
                  Jianwei Zhang and
                  Jian Yang and
                  Jiaxi Yang and
                  Jingren Zhou and
                  Junyang Lin and
                  Kai Dang and
                  Keqin Bao and
                  Kexin Yang and
                  Le Yu and
                  Lianghao Deng and
                  Mei Li and
                  Mingfeng Xue and
                  Mingze Li and
                  Pei Zhang and
                  Peng Wang and
                  Qin Zhu and
                  Rui Men and
                  Ruize Gao and
                  Shixuan Liu and
                  Shuang Luo and
                  Tianhao Li and
                  Tianyi Tang and
                  Wenbiao Yin and
                  Xingzhang Ren and
                  Xinyu Wang and
                  Xinyu Zhang and
                  Xuancheng Ren and
                  Yang Fan and
                  Yang Su and
                  Yichang Zhang and
                  Yinger Zhang and
                  Yu Wan and
                  Yuqiong Liu and
                  Zekun Wang and
                  Zeyu Cui and
                  Zhenru Zhang and
                  Zhipeng Zhou and
                  Zihan Qiu},
  title        = {Qwen3 Technical Report},
  journal      = {CoRR},
  volume       = {abs/2505.09388},
  year         = {2025},
  url          = {https://doi.org/10.48550/arXiv.2505.09388},
  doi          = {10.48550/ARXIV.2505.09388},
  eprinttype    = {arXiv},
  eprint       = {2505.09388},
  bibsource    = {dblp computer science bibliography, https://dblp.org}
}

@article{jiang2023mistral7b,
      author       = {Albert Q. Jiang and
                  Alexandre Sablayrolles and
                  Arthur Mensch and
                  Chris Bamford and
                  Devendra Singh Chaplot and
                  Diego de Las Casas and
                  Florian Bressand and
                  Gianna Lengyel and
                  Guillaume Lample and
                  Lucile Saulnier and
                  L{\'{e}}lio Renard Lavaud and
                  Marie{-}Anne Lachaux and
                  Pierre Stock and
                  Teven Le Scao and
                  Thibaut Lavril and
                  Thomas Wang and
                  Timoth{\'{e}}e Lacroix and
                  William El Sayed},
  title        = {Mistral 7B},
  journal      = {CoRR},
  volume       = {abs/2310.06825},
  year         = {2023},
  url          = {https://doi.org/10.48550/arXiv.2310.06825},
  doi          = {10.48550/ARXIV.2310.06825},
  eprinttype    = {arXiv},
  eprint       = {2310.06825},
  bibsource    = {dblp computer science bibliography, https://dblp.org} 
}

@inproceedings{jiang2025ragboostefficientretrievalaugmentedgeneration,
      title={RAGBoost: Efficient Retrieval-Augmented Generation with Accuracy-Preserving Context Reuse}, 
      author={Yinsicheng Jiang and Yeqi Huang and Liang Cheng and Cheng Deng and Xuan Sun and Luo Mai},
      year={2025},
      eprint={2511.03475},
      archivePrefix={arXiv},
      primaryClass={cs.LG},
      url={https://arxiv.org/abs/2511.03475}, 
}

@inproceedings{hu2021loralowrankadaptationlarge,
      author       = {Edward J. Hu and
                  Yelong Shen and
                  Phillip Wallis and
                  Zeyuan Allen{-}Zhu and
                  Yuanzhi Li and
                  Shean Wang and
                  Lu Wang and
                  Weizhu Chen},
  title        = {LoRA: Low-Rank Adaptation of Large Language Models},
  booktitle    = {The Tenth International Conference on Learning Representations, {ICLR}
                  2022, Virtual Event, April 25-29, 2022},
  publisher    = {OpenReview.net},
  year         = {2022},
  url          = {https://openreview.net/forum?id=nZeVKeeFYf9},
  bibsource    = {dblp computer science bibliography, https://dblp.org} 
}

@inprpceedings{zheng2024sglangefficientexecutionstructured,
      author       = {Lianmin Zheng and
                  Liangsheng Yin and
                  Zhiqiang Xie and
                  Chuyue Sun and
                  Jeff Huang and
                  Cody Hao Yu and
                  Shiyi Cao and
                  Christos Kozyrakis and
                  Ion Stoica and
                  Joseph E. Gonzalez and
                  Clark W. Barrett and
                  Ying Sheng},
  editor       = {Amir Globersons and
                  Lester Mackey and
                  Danielle Belgrave and
                  Angela Fan and
                  Ulrich Paquet and
                  Jakub M. Tomczak and
                  Cheng Zhang},
  title        = {SGLang: Efficient Execution of Structured Language Model Programs},
  booktitle    = {Advances in Neural Information Processing Systems 38: Annual Conference
                  on Neural Information Processing Systems 2024, NeurIPS 2024, Vancouver,
                  BC, Canada, December 10 - 15, 2024},
  year         = {2024},
  url          = {http://papers.nips.cc/paper\_files/paper/2024/hash/724be4472168f31ba1c9ac630f15dec8-Abstract-Conference.html},
  bibsource    = {dblp computer science bibliography, https://dblp.org}
}

@article{qin2025mooncakekvcachecentricdisaggregatedarchitecture,
      author       = {Ruoyu Qin and
                  Zheming Li and
                  Weiran He and
                  Mingxing Zhang and
                  Yongwei Wu and
                  Weimin Zheng and
                  Xinran Xu},
  title        = {Mooncake: {A} KVCache-centric Disaggregated Architecture for {LLM}
                  Serving},
  journal      = {CoRR},
  volume       = {abs/2407.00079},
  year         = {2024},
  url          = {https://doi.org/10.48550/arXiv.2407.00079},
  doi          = {10.48550/ARXIV.2407.00079},
  eprinttype    = {arXiv},
  eprint       = {2407.00079},
  bibsource    = {dblp computer science bibliography, https://dblp.org}
}

@inproceedings{zhang2025attentionentropykeyfactor,
      author       = {Zhisong Zhang and
                  Yan Wang and
                  Xinting Huang and
                  Tianqing Fang and
                  Hongming Zhang and
                  Chenlong Deng and
                  Shuaiyi Li and
                  Dong Yu},
  editor       = {Wanxiang Che and
                  Joyce Nabende and
                  Ekaterina Shutova and
                  Mohammad Taher Pilehvar},
  title        = {Attention Entropy is a Key Factor: An Analysis of Parallel Context
                  Encoding with Full-attention-based Pre-trained Language Models},
  booktitle    = {Proceedings of the 63rd Annual Meeting of the Association for Computational
                  Linguistics (Volume 1: Long Papers), {ACL} 2025, Vienna, Austria,
                  July 27 - August 1, 2025},
  pages        = {9840--9855},
  publisher    = {Association for Computational Linguistics},
  year         = {2025},
  url          = {https://aclanthology.org/2025.acl-long.485/},
  bibsource    = {dblp computer science bibliography, https://dblp.org} 
}

@article{su2023roformerenhancedtransformerrotary,
      author       = {Jianlin Su and
                  Murtadha H. M. Ahmed and
                  Yu Lu and
                  Shengfeng Pan and
                  Wen Bo and
                  Yunfeng Liu},
  title        = {RoFormer: Enhanced transformer with Rotary Position Embedding},
  journal      = {Neurocomputing},
  volume       = {568},
  pages        = {127063},
  year         = {2024},
  url          = {https://doi.org/10.1016/j.neucom.2023.127063},
  doi          = {10.1016/J.NEUCOM.2023.127063},
  bibsource    = {dblp computer science bibliography, https://dblp.org}
}

@misc{xiao2024efficientstreaminglanguagemodels,
      author       = {Guangxuan Xiao and
                  Yuandong Tian and
                  Beidi Chen and
                  Song Han and
                  Mike Lewis},
  title        = {Efficient Streaming Language Models with Attention Sinks},
  booktitle    = {The Twelfth International Conference on Learning Representations,
                  {ICLR} 2024, Vienna, Austria, May 7-11, 2024},
  publisher    = {OpenReview.net},
  year         = {2024},
  url          = {https://openreview.net/forum?id=NG7sS51zVF},
  bibsource    = {dblp computer science bibliography, https://dblp.org}
}

@misc{ragboost,
      title={RAGBoost: Efficient Retrieval-Augmented Generation with Accuracy-Preserving Context Reuse}, 
      author={Yinsicheng Jiang and Yeqi Huang and Liang Cheng and Cheng Deng and Xuan Sun and Luo Mai},
      year={2025},
      eprint={2511.03475},
      archivePrefix={arXiv},
      primaryClass={cs.LG},
      url={https://arxiv.org/abs/2511.03475}, 
}

@inproceedings{metis,
  author       = {Siddhant Ray and
                  Rui Pan and
                  Zhuohan Gu and
                  Kuntai Du and
                  Shaoting Feng and
                  Ganesh Ananthanarayanan and
                  Ravi Netravali and
                  Junchen Jiang},
  editor       = {Youjip Won and
                  Youngjin Kwon and
                  Ding Yuan and
                  Rebecca Isaacs},
  title        = {{METIS:} Fast Quality-Aware {RAG} Systems with Configuration Adaptation},
  booktitle    = {Proceedings of the {ACM} {SIGOPS} 31st Symposium on Operating Systems
                  Principles, {SOSP} 2025, Lotte Hotel World, Seoul, Republic of Korea,
                  October 13-16, 2025},
  pages        = {606--622},
  publisher    = {{ACM}},
  year         = {2025},
  url          = {https://doi.org/10.1145/3731569.3764855},
  doi          = {10.1145/3731569.3764855},
  bibsource    = {dblp computer science bibliography, https://dblp.org}
}

@inproceedings{hydrarag,
  author       = {Zhengding Hu and
                  Vibha Murthy and
                  Zaifeng Pan and
                  Wanlu Li and
                  Xiaoyi Fang and
                  Yufei Ding and
                  Yuke Wang},
  editor       = {Youjip Won and
                  Youngjin Kwon and
                  Ding Yuan and
                  Rebecca Isaacs},
  title        = {HedraRAG: Co-Optimizing Generation and Retrieval for Heterogeneous
                  {RAG} Workflows},
  booktitle    = {Proceedings of the {ACM} {SIGOPS} 31st Symposium on Operating Systems
                  Principles, {SOSP} 2025, Lotte Hotel World, Seoul, Republic of Korea,
                  October 13-16, 2025},
  pages        = {623--638},
  publisher    = {{ACM}},
  year         = {2025},
  url          = {https://doi.org/10.1145/3731569.3764806},
  doi          = {10.1145/3731569.3764806},
  bibsource    = {dblp computer science bibliography, https://dblp.org}
}

@misc{smallvslarge,
      title={Train Large, Then Compress: Rethinking Model Size for Efficient Training and Inference of Transformers}, 
      author={Zhuohan Li and Eric Wallace and Sheng Shen and Kevin Lin and Kurt Keutzer and Dan Klein and Joseph E. Gonzalez},
      year={2020},
      eprint={2002.11794},
      archivePrefix={arXiv},
      primaryClass={cs.CL},
      url={https://arxiv.org/abs/2002.11794}, 
}

@book{f1,
  author       = {C. J. van Rijsbergen},
  title        = {Information Retrieval},
  publisher    = {Butterworth},
  year         = {1979},
  isbn         = {0-408-70929-4},
  bibsource    = {dblp computer science bibliography, https://dblp.org}
}

@String{Computing = "Computing" }

@String{Computer = "{IEEE} Computer" }

\end{document}